\pdfoutput=1

\documentclass[11pt]{article}

\usepackage{acl}

\hypersetup{pdfauthor={Yaron Anavi, Mor Aisenberg, Nadav Nesher, Elena Khabibullina, Isabella Cattinelli}, pdftitle={Knowing When to Stop: Predicting Execution-Consistency Convergence in Text-to-SQL}}

\usepackage{times}
\usepackage{latexsym}
\usepackage[T1]{fontenc}
\usepackage[utf8]{inputenc}
\usepackage{microtype}
\usepackage{inconsolata}

\usepackage{amsmath,amssymb,amsfonts}
\usepackage{graphicx}
\usepackage{booktabs}
\usepackage[table]{xcolor}
\usepackage{multirow}
\usepackage[section]{placeins}
\usepackage{float}

\title{Knowing When to Stop: Predicting Execution-Consistency Convergence in Text-to-SQL}

\author{
  Yaron Anavi, \ Mor Aisenberg, \ Nadav Nesher, \ Elena Khabibullina, \ Isabella Cattinelli \\[6pt]
  {\normalfont GIGASPACES} \\
}


\begin{document}

\maketitle

\begin{abstract}
Repeated LLM calls are the standard way to estimate how trustworthy a Text-to-SQL result is: run the pipeline multiple times, judge each SQL execution, and use the consistency of the verdicts as a confidence signal. The open question is \emph{when to stop, when the consistency has converged}. We formulate this as a convergence-prediction problem and train a family of lightweight 1-D models that observe the running consistency trajectory and decide, at each step, whether further runs are unlikely to shift it materially, and we benchmark them against a principled Beta-Bernoulli stopping rule and a learned run-count baseline. On the BIRD benchmark and two production customer datasets, our method adapts its stopping point to each user question, halting sooner when consistency converges early and continuing longer when it converges late. We further show that the weak serial correlation between runs lets us permute their order as a training augmentation, controlled by a tunable shuffling weight. Performance stays consistent across the three datasets, and to mimic an imperfect production judge we inject noise into the correct/incorrect verdicts obtained by comparing the generated and ground-truth SQL results, showing that the method still predicts convergence reliably.
\end{abstract}

\section{Introduction}
\label{sec:introduction}

\begin{figure}[t]
\centering
\includegraphics[width=\columnwidth]{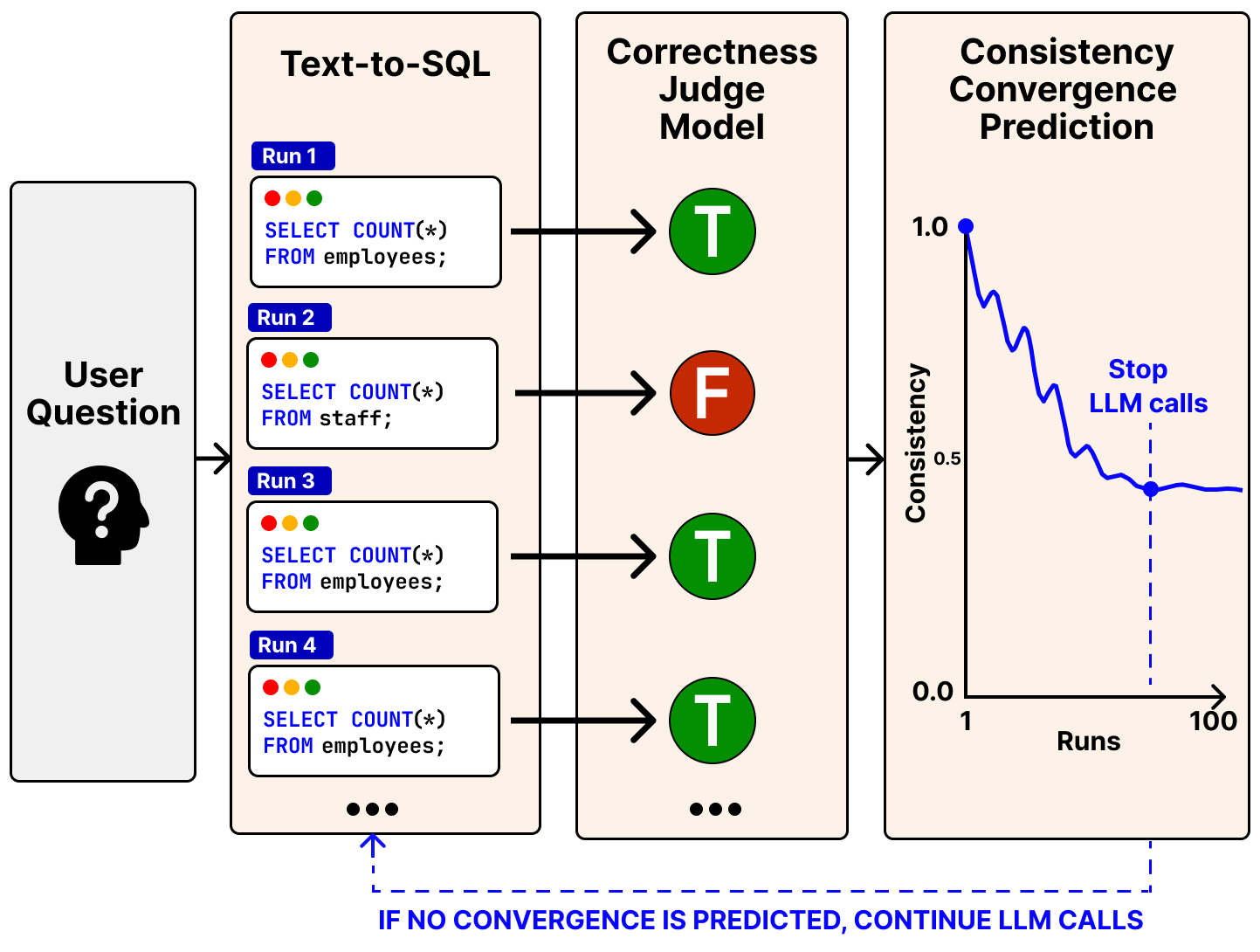}
\caption{Stopping-criterion pipeline. A user question is repeatedly run by an LLM that produces SQL via Text-to-SQL; each generated SQL's correctness is judged by examining the produced SQL and its execution result. This produces a sequence of binary outcomes, transposed into a 1-D consistency signal. We propose a framework in which, once a correctness judge is plugged in, a lightweight model predicts when the sequence of judge verdicts has converged.}
\label{fig:architecture}
\end{figure}

Modern production Text-to-SQL pipelines translate a user question into SQL, execute it, and check whether its execution result matches the expected answer~\citep{li2024bird, lei2024spider2}. Because large language models can respond inconsistently to the same prompt, a single run is an unreliable witness, so pipelines run the same question several times and use the \emph{consistency} of the outcomes as a trust signal: high agreement suggests the answer can be trusted, while low agreement suggests the model is uncertain or the question is ambiguous and may need a clarifying follow-up. The stakes are higher for Text-to-SQL than for free-form text: a wrong column, a missing join condition, or an off-by-one filter produces an entirely wrong result table rather than a gracefully degraded answer.

This reliability comes at a cost. The more runs we collect, the more we can trust the consistency estimate, but each run is a full generation plus execution, which in production translates directly into latency and budget~\citep{aggarwal2023adaptive, li2024esc}. Fixed-budget stopping, running every question a preset number of times, wastes calls on easy questions that settle immediately and starves hard ones that need more. The central question is therefore: \emph{when has the consistency level converged enough that further runs are unlikely to change it?} Answering it turns consistency estimation from a fixed-budget procedure into an adaptive one.

We study this question on both benchmark and production customer datasets. For each question we hold an expected (ground-truth) SQL alongside the model-generated SQL; executing both and comparing their result tables turns every run into a binary \texttt{True}/\texttt{False} outcome. From the running sequence of these outcomes we learn the run at which the consistency converges, so the pipeline can stop there and read off the consistency reached just before convergence. In a real production system the expected SQL does not exist; instead one plugs in a \emph{judge} meant to approximate this comparison as closely as possible, that is, to decide whether the generated SQL produces the desired table. To test whether our method would survive such a judge, we inject label noise that simulates an imperfect one and ask whether convergence is still predictable. If plugging in a judge of Text-to-SQL correctness in place of the ground-truth comparison still lets us predict convergence, then such a judge could be dropped into production.

\paragraph{Contributions.}
\begin{itemize}
    \item We present a method that allocates the right run budget for judging Text-to-SQL correctness: it reads the running consistency as a 1-D signal and outputs both when to stop and the consistency at that point (Sections~\ref{sec:problem_setup} and~\ref{sec:methodology}).
    \item We train a family of lightweight convergence predictors that, at each LLM run and SQL execution, classify whether the consistency has converged and detect the first converged run, outperforming both a fixed-budget rule that runs every question a preset number of times and a principled Beta-Bernoulli stopping rule (Section~\ref{sec:experiments_and_results}).
    \item We stress-test the method by injecting label noise to simulate an imperfect judge and show it degrades gracefully, supporting its use as a drop-in stopping layer above any binary execution judge.
    \item We find the serial correlation between nearby runs is weak, justifying run-order permutation as a training augmentation, controlled by a tunable shuffling weight (Sections~\ref{sec:problem_setup} and~\ref{sec:methodology}).
\end{itemize}

\section{Related Work}
\label{sec:related_work}

Our work draws on four threads: judging LLM responses, Text-to-SQL generation, consistency in repeated LLM outputs, and adaptive stopping for LLM inference.

\paragraph{Judging LLM responses.}
Using one LLM to judge or \emph{critique} another's output is now a large, active area, spanning general-purpose evaluators~\citep{liu2023geval, zheng2023judging} and critic models that propose corrections~\citep{gou2024critic}. The same idea is increasingly applied to Text-to-SQL, where learned judges decide execution correctness or semantic \emph{equivalence}~\citep{kim-etal-2025-flex} and actor--critic or clause-wise critics detect and repair errors~\citep{zheng2024actorcritic, chen2025sqlcritic, askari2024magic}. We use a lab-controlled, ground-truth judge as a clean stand-in; in production, exactly these learned judge models are what would plug in above our pipeline.

\paragraph{Text-to-SQL generation.}
Generating SQL from natural language is a large field of its own, from sequence-to-sequence models~\citep{zhong2017seq2sql} and cross-domain and enterprise benchmarks~\citep{yu2018spider, lei2024spider2} to modern LLM pipelines built on schema linking, decomposition, and self-correction~\citep{pourreza2023dinsql, pourreza2025chasesql}, and to large-scale benchmarking of such methods~\citep{gao2024dailsql}. Our experiments use the financial-domain subset of the BIRD benchmark~\citep{li2024bird}, and our pipeline takes the per-run correctness outcome these systems produce as input rather than contributing to generation itself.

\paragraph{Consistency in repeated LLM outputs.}
Self-Consistency~\citep{wang2023selfconsistency} samples multiple chain-of-thought paths and takes the majority answer; Universal Self-Consistency~\citep{chen2023universal} extends this to open-ended generation, and \citet{brown2024monkeys} show that repeated sampling improves answer coverage on hard reasoning benchmarks. Consistency has also been linked to calibration: \citet{xiong2024llmuncertainty} find behavioral consistency more calibrated than verbalized confidence, and \citet{kuhn2023semantic} propose semantic uncertainty for generation. We adopt majority-vote consistency as our signal.

\paragraph{Adaptive stopping criteria for LLM inference.}
Adaptive-Consistency~\citep{aggarwal2023adaptive} places Dirichlet/beta-binomial posteriors over the running answer tallies and stops once the majority lead is statistically secure, cutting the sample budget sharply at negligible accuracy cost; we instantiate its binary case as a Beta-Bernoulli posterior stopping rule and use it as our principled statistical baseline (Section~\ref{sec:experiments_and_results}). Early-Stopping Self-Consistency~\citep{li2024esc} is a simpler, parameter-light rule that halts once a window of samples agrees. Closest to our setting, \citet{qu2025betaconform} model an LLM-ensemble judge's outputs with a mixture of Beta--Binomial distributions and apply conformal-prediction-based adaptive stopping during sampling; we instead predict per-question execution-consistency convergence with a learned signal model. Relatedly, \citet{chen2023frugalgpt} cascade models by response confidence to cut inference cost.

\section{Problem Setup}
\label{sec:problem_setup}

We focus on a specific notion of consistency: agreement of the \emph{post-process evaluation} of Text-to-SQL responses. Each LLM response is turned into SQL through chain-of-thought reasoning guided by prompt rules, executed against the database, and labeled \texttt{True} when the resulting table matches the expected table exactly or within a similarity threshold (Section~\ref{sec:methodology}) and \texttt{False} otherwise. We study this in a controlled lab setting: each label is created by comparing the generated table against the dataset's ground-truth expected table, with no learned judge or critic model in the loop. Collecting these labels across $n$ runs of the same question, we define the consistency as
\begin{equation}
    \text{Consistency}(n) = \frac{\max\bigl(N_{\texttt{True}}(n),\; N_{\texttt{False}}(n)\bigr)}{n},
    \label{eq:consistency}
\end{equation}
where $N_{\texttt{True}}(n)$ and $N_{\texttt{False}}(n)$ denote the counts of the two outcome classes. Figure~\ref{fig:convergence_examples} (Appendix~\ref{sec:experimental_setup}) shows how this value settles as runs accumulate. Because it takes the larger of the two counts, a question whose runs are all \texttt{False}, the generated and expected tables never match, is $100\%$ consistent even though the Text-to-SQL output is uniformly wrong; we are measuring the consistency of the Text-to-SQL behavior, not necessarily its success.

\paragraph{Convergence criterion.}
For a prefix of $n$ binary responses, we define the consistency as \emph{converged} at run $n$ if it stays within $K$ percentage points of $\text{C}(n)$ over the next $W$ runs:
\begin{equation}
    \resizebox{0.92\columnwidth}{!}{$%
    \text{Converged}(n, W, K) = \mathbf{1}\!\left[\max\limits_{n < t \leq n+W} \bigl|\text{C}(t) - \text{C}(n)\bigr| \leq K\right]%
    $}
    \label{eq:stability}
\end{equation}
where $\mathbf{1}[\cdot]$ equals $1$ when the bracketed condition holds and $0$ otherwise. Throughout this work we set $W = 30$ and $K = 0.05$. We chose a window-based criterion because it is simple yet effective: in our datasets, once the consistency stays within $K$ percentage points over a $30$-run window, it remains within $K$ for the rest of the runs with empirical probability $0.995$. The fixed $W$-run window also gives a concrete way to evaluate convergence: deciding whether run $n$ has converged requires the next $W$ runs as a verification window, so convergence is hard to evaluate once too few runs remain to fill it.

\paragraph{Are runs i.i.d., and what does this imply?}
Whether repeated runs are independent shapes how we model them, in particular whether to account for temporal correlation. Recent work shows they are not fully i.i.d.: even at temperature $0$, the same prompt can produce different outputs depending on serving conditions such as batch size and load~\citep{he2025nondeterminism}. This matters because, if we model each run's \texttt{True}/\texttt{False} outcome as i.i.d.\ Bernoulli with success probability $p$, the running consistency is the empirical mean $\bar{X}_n = S_n/n$ (where $S_n$ counts the \texttt{True} outcomes in the first $n$ runs), whose variance is
\begin{equation}
    \mathrm{Var}(\bar{X}_n) = \frac{p(1-p)}{n},
    \label{eq:variance}
\end{equation}
Since this variance is highly correlated with convergence and shrinks as $p$ grows, the running consistency at run $n$ is itself a good feature: it predicts how quickly the variance will become small, which in turn signals convergence. The lag-$k$ autocorrelation of each question's outcome sequence ($k=1,2,3$) stays near zero across datasets (Table~\ref{tab:autocorr}), so the i.i.d.\ assumption is a reasonable approximation.

\begin{table}[t]
    \centering
    \caption{Lag-$k$ autocorrelation of the per-question binary Text-to-SQL correctness sequence (the \texttt{True}/\texttt{False} outcomes), averaged across questions and datasets. Observed values stay within $\pm 0.05$ of zero, indicating only weak serial dependence.}
    \label{tab:autocorr}
    \footnotesize
    \setlength{\tabcolsep}{4pt}
    \begin{tabular}{lccc}
        \toprule
        \textbf{Dataset} & \textbf{lag-1} & \textbf{lag-2} & \textbf{lag-3} \\
        \midrule
        BIRD & $-0.0023$ & $-0.0143$ & $-0.0380$ \\
        Dataset A    & $-0.0333$ & $-0.0092$ & $\phantom{-}0.0015$ \\
        Dataset B    & $\phantom{-}0.0198$ & $-0.0073$ & $-0.0009$ \\
        \bottomrule
    \end{tabular}
\end{table}

\section{Methodology}
\label{sec:methodology}

We treat the running consistency as a 1-D signal and predict when it has converged, combining two input \emph{schemes} with two task \emph{variants}, evaluated both under clean ground-truth labels and under injected label noise (Section~\ref{sec:experiments_and_results}). The injected noise flips each \texttt{True}/\texttt{False} comparison outcome with some probability, a realistic and fairly close approximation of an imperfect judge, though still not an exact model of a real one, whose errors need not be random.

\subsection{Inputs and Tasks}
\label{subsec:formulation}

Let $\text{con}(n)$ denote the consistency after $n$ runs (Equation~\ref{eq:consistency}). We feed the sequence $\bigl(\text{con}(1), \ldots, \text{con}(n)\bigr)$ to the model in two ways:
\begin{itemize}
    \item \textbf{Feature-based scheme:} a hand-crafted feature vector $\mathbf{m}_n \in \mathbb{R}^d$ summarizing the trajectory up to run $n$. We use $d = 6$ features: (i)~the run index $n$; (ii)~the current consistency $\text{con}(n)$; and (iii)~the local sample variance $\sigma^2_w(n) = \frac{1}{w}\sum_{t=n-w+1}^{n} \bigl(\text{con}(t) - \bar{\text{con}}_w(n)\bigr)^2$ of the consistency over the most recent $w$ runs for $w \in \{5, 10, 15, 30\}$, where $\bar{\text{con}}_w(n)$ is the window mean. When $n < w$, we use the runs available so far.
    \item \textbf{Raw-signal scheme:} the raw consistency trajectory $\bigl(\text{con}(1), \ldots, \text{con}(n)\bigr)$ fed directly to a 1-D deep model.
\end{itemize}

On top of either scheme, we train two task variants. The first is \textbf{classification}: predict, at each run $n$, the binary label $y_n = \text{Converged}(n, 30, 0.05)$ from Equation~\ref{eq:stability}, evaluated with ROC AUC. The second is \textbf{detection}: \textbf{we reuse the same models but tune them for this task}, which classify, at any given run, whether convergence has been reached, and stop at the first run they flag as converged, optimized to fire as close as possible to the \emph{first} converged run, evaluated with RMSE between the detected run $n_{\text{detected}}$ and the true first converged run $n_{\text{first\_conv}}$:
\begin{equation}
    \mathcal{L}_{\text{detect}} = \sqrt{\frac{1}{N}\sum_{i=1}^{N} \bigl(n_{\text{detected}}^{(i)} - n_{\text{first\_conv}}^{(i)}\bigr)^2}.
    \label{eq:detection_loss}
\end{equation}
Here $N$ is the number of test examples. Because a premature stop returns an unconverged estimate, we bias the detector toward later stopping: although the natural classification threshold is $0.5$, we tune the decision threshold on the validation set toward later detection, so the method is biased against firing too early. Figure~\ref{fig:architecture} illustrates the per-step decision pipeline.

\subsection{Data Augmentation via Permutation}
\label{subsec:augmentation}

Since runs exhibit weak temporal dependence (Table~\ref{tab:autocorr}), we train on a mix of the natural run order and permuted variants. A hyperparameter $\beta_{\text{shuffle}}$ weights the original order against the permutations, tuned per dataset during training. At test time the run order is never permuted, so evaluation reflects the actual sequential setting.

\subsection{Models}
\label{subsec:models}

For the feature-based scheme we use \emph{XGBoost}, which is more expressive, and \emph{logistic regression}, which is simpler and harder to overfit on our modest data. For the raw-signal scheme we use a small 1-D \emph{temporal convolutional network} (TCN)~\citep{bai2018tcn} that learns its own representation through dilated 1-D convolutions. With more data, we expect deeper and more complex models, such as transformer-based 1-D networks like PatchTST~\citep{nie2023patchtst}, to be a better fit.

\section{Experiments and Results}
\label{sec:experiments_and_results}

\subsection{Datasets}
\label{subsec:datasets}

Our dataset consists of multiple user-asked questions, each associated with multiple runs of the same user question. Each run is assigned a binary outcome, \texttt{True} or \texttt{False}, based on whether the generated SQL produces a result matching that of the manually labeled expected SQL (Figure~\ref{fig:pipeline}). Our Text-to-SQL system is inspired by BIRD~\citep{li2024bird} and uses chain-of-thought reasoning over the database schema, business rules, target SQL dialect, and few-shot similar-question patterns. All generations were produced with Azure OpenAI GPT-4.1~\citep{openai2025gpt41}.

We evaluate on three datasets (Table~\ref{tab:dataset_stats}). The first is the finance-domain subset of the public BIRD benchmark~\citep{li2024bird}, which we refer to as BIRD throughout. The other two are anonymized customer datasets from the solar-energy and outdoor-products manufacturing domains, used with the data providers' permission. The three differ in number of questions ($83$, $25$, $19$) and mean Text-to-SQL consistency ($\sim$84\%--95\%), where a question's consistency is the majority fraction over its $100$ runs, the larger share of runs agreeing on whether the generated SQL's result matches that of the expected SQL.

\begin{table}[!htb]
    \centering
    \caption{Summary statistics for the three evaluation datasets. The BIRD row corresponds to a subset of the public BIRD benchmark; the remaining two are anonymized customer datasets. All use 100 runs per question.}
    \label{tab:dataset_stats}
    \footnotesize
    \setlength{\tabcolsep}{4pt}
    \begin{tabular}{lccc}
        \toprule
        \textbf{Dataset} & \textbf{Questions} & \textbf{Max runs} & \textbf{Mean consistency (\%)} \\
        \midrule
        BIRD & 83 & 100 & 95.30 \\
        Dataset A    & 25 & 100 & 84.08 \\
        Dataset B    & 19 & 100 & 93.53 \\
        \bottomrule
    \end{tabular}
\end{table}

\paragraph{Ground-truth labeling.} For each run, the binary outcome \texttt{True}/\texttt{False} comes from comparing the table executed from the generated SQL against the table from the manually annotated expected SQL. The rule is tailored to our product needs but can be tightened or loosened per deployment:
\begin{itemize}
    \item Identical tables (same rows, columns, values) are labeled \texttt{True}.
    \item Otherwise, the run is labeled \texttt{True} if (i) the expected columns are a subset of the generated columns (so all expected information is present, possibly alongside extras) and (ii) the subset tables extracted from the shared columns are identical.
\end{itemize}
Figure~\ref{fig:pipeline} (Appendix~\ref{sec:experimental_setup}) illustrates the labeling pipeline.

\subsection{Classification Scheme: Convergence Prediction}
\label{subsec:classification_scheme}

We evaluate every model with strict question-level $5$-fold cross-validation: each fold holds out the data extracted from $\sim$20\% of questions for test, and the data from the remaining four folds is split into $\sim$70\% train and $\sim$10\% validation (used for early stopping and hyperparameter selection). All metrics are averaged over the five folds.

\paragraph{Training data.} For each question, we slide over runs and form examples of the prefix up to $n$ (as feature vector or raw consistency sequence) paired with the label $y_n = \text{Converged}(n, 30, 0.05)$ from Equation~\ref{eq:stability}. To leave room for the $30$-run verification window, we restrict $n \leq 70$. For instance, in BIRD this extracts $70$ prefixes from each of its $83$ questions, a binary-classification dataset of $70 \times 83 = 5810$ examples (unshuffled; the shuffled regime adds $10\times$ more, the optimum from our grid search).

\paragraph{Per-dataset training.} A separate model is trained per dataset, reflecting realistic deployment and serving as a generalization check across the three domains.

\paragraph{Models.} XGBoost, logistic regression, and a 1-D TCN, together with the Beta-Bernoulli stopping rule as a principled statistical baseline. For each learned model, hyperparameters are selected by grid search on the validation fold.

\paragraph{Shuffled vs.\ unshuffled.} In the \emph{shuffled} regime we add $10\times$ permuted prefixes per question, random permutations of the original run order, fixed once to keep the comparison strict, with $\beta_{\text{shuffle}}$ weighting the original ordering against the permutations. The \emph{unshuffled} regime uses only the natural order. Test order is never permuted.

\paragraph{Baselines.} We compare against two baselines. The first, \emph{runs-only}, is the de-facto fixed-budget rule used in production. The second is a \emph{Beta-Bernoulli} stopping rule, the binary-case reduction of Adaptive-Consistency~\citep{aggarwal2023adaptive}: after each run we form the posterior $\mathrm{Beta}(1{+}N_{\texttt{True}},\, 1{+}N_{\texttt{False}})$ over the success probability and stop once its credible-interval width falls below a threshold. Its parameters, the interval level and width threshold, are tuned on the validation fold, as with the learned models. For completeness we also evaluate single-feature inputs (consistency only, variation only).

\subsection{Classification: AUC across Configurations}
\label{subsec:results_classification}

Table~\ref{tab:auc_noise_results} reports ROC AUC averaged across the three datasets and five folds. Shuffled augmentation improves every model, dramatically for the raw-signal TCN (from $0.760$ to $0.881$) and marginally for the feature-based models. \textbf{XGBoost with the full feature set is best on clean labels at $0.913$ AUC}, with logistic regression ($0.908$) and the TCN ($0.881$) close behind. Under label noise the models degrade gracefully: XGBoost holds at $0.887$ at $5\%$ flips and $0.878$ at $15\%$, and the feature-based models stay above $0.85$ even at $15\%$. All comfortably outperform the Beta-Bernoulli (ASC) baseline, which reaches $0.876$ on clean labels and falls to $0.808$ at $15\%$.

\begin{table}[!htb]
    \centering
    \caption{ROC AUC for convergence prediction, averaged across three datasets and five folds. Headline configurations per model family, the run-count baseline, and the Beta-Bernoulli (Adaptive-Consistency, ASC) stopping rule. The full single-feature ablation is in Appendix~\ref{sec:experimental_setup} (Table~\ref{tab:auc_full}). ``Shuf.'' indicates permutation-based augmentation; the $5\%$ and $15\%$ noise columns randomly flip that fraction of the \texttt{True}/\texttt{False} labels.}
    \label{tab:auc_noise_results}
    \footnotesize
    \setlength{\tabcolsep}{3pt}
    \begin{tabular}{llcccc}
        \toprule
        \textbf{Method} & \textbf{Features} & \textbf{Shuf.} & \textbf{clean} & \textbf{5\%} & \textbf{15\%} \\
        \midrule
        Logistic & runs+cons.+var. & yes & 0.908 & 0.876 & 0.870 \\
        Logistic & runs+cons.+var. & no  & 0.905 & 0.867 & 0.861 \\
        Logistic & runs only       & yes & 0.744 & 0.836 & 0.859 \\
        Logistic & runs only       & no  & 0.735 & 0.835 & 0.857 \\
        \midrule
        \rowcolor{gray!20} XGBoost & runs+cons.+var. & yes & 0.913 & 0.887 & 0.878 \\
        XGBoost & runs+cons.+var. & no  & 0.899 & 0.882 & 0.856 \\
        \midrule
        TCN & raw signal & yes & 0.881 & 0.862 & 0.861 \\
        TCN & raw signal & no  & 0.760 & 0.837 & 0.860 \\
        \midrule
        ASC & Beta posterior & --- & 0.876 & 0.848 & 0.808 \\
        \bottomrule
    \end{tabular}
\end{table}

\subsection{Detection Scheme}
\label{subsec:detection_scheme}

The detection scheme reuses the classification scheme's data construction, splits, model families, hyperparameter grid, and shuffled/unshuffled variants. The objective is what changes: detection predicts \emph{when} convergence first happens. Working at the level of the whole question rather than individual run prefixes, we compare the model's first-firing run $n_{\text{detected}}$ to the true first converged run $n_{\text{first\_conv}}$, then select the configuration minimizing the RMSE between them (Equation~\ref{eq:detection_loss}), reported as an average across folds and models. Hyperparameters here are tuned to place the stopping run correctly, minimizing this run-level RMSE, rather than to maximize accuracy at each individual run. The decision threshold, fixed at $0.5$ for classification, is itself tuned in this setting, controlling how sensitive detection is and hence whether it fires too early or too late.

\subsection{Detection: RMSE across Configurations}
\label{subsec:results_detection}

Table~\ref{tab:rmse_noise_results} reports detection RMSE across model families. \textbf{Logistic regression with the full feature set is best at RMSE $8.18$}, ahead of XGBoost, the TCN, and the runs-only baseline. It is resilient to noise, holding at $8.21$ at $5\%$ flips and $8.30$ at $15\%$. Shuffled augmentation again helps across models, with its clearest payoff on detection (logistic: $8.18$ vs.\ $9.43$ unshuffled). All outperform the Beta-Bernoulli (ASC) baseline, which trails at $12.82$ and degrades further under noise. Our method also reduces total LLM calls: summed across all questions, it uses $35\%$ fewer than the Beta-Bernoulli baseline. If we focus on the questions that converge very fast, within the first $10$ runs ($21\%$ of the dataset), our method saves $42\%$ of runs compared to the Beta-Bernoulli baseline and is also more accurate over that interval (RMSE $\approx 5$ vs.\ $\approx 12$ runs). These call savings are not uniform across questions: the method reduces calls when consistency converges early but may use more than the baseline when it converges late, where the additional calls yield a more accurate consistency estimate. We expect this adaptivity to matter most on datasets with large variation in convergence points, where no single fixed budget fits all questions well.

\begin{table}[!htb]
    \centering
    \caption{Detection RMSE between $n_{\text{detected}}$ and the true first converged run $n_{\text{first\_conv}}$, averaged across three datasets and five folds (lower is better). Headline configurations per family, the run-count baseline, and the Beta-Bernoulli (ASC) rule; the full single-feature ablation is in Appendix~\ref{sec:experimental_setup} (Table~\ref{tab:rmse_full}). ``Shuf.'' indicates permutation-based augmentation; the $5\%$ and $15\%$ noise columns randomly flip that fraction of the \texttt{True}/\texttt{False} labels.}
    \label{tab:rmse_noise_results}
    \footnotesize
    \setlength{\tabcolsep}{3pt}
    \begin{tabular}{llcccc}
        \toprule
        \textbf{Method} & \textbf{Features} & \textbf{Shuf.} & \textbf{clean} & \textbf{5\%} & \textbf{15\%} \\
        \midrule
        \rowcolor{gray!20} Logistic & runs+cons.+var. & yes & \phantom{0}8.18 & \phantom{0}8.21 & \phantom{0}8.30 \\
        Logistic & runs+cons.+var. & no  & \phantom{0}9.43 & \phantom{0}8.98 & \phantom{0}8.59 \\
        Logistic & runs only       & yes & 12.25 & 11.49 & \phantom{0}9.70 \\
        Logistic & runs only       & no  & 14.22 & 13.14 & 10.02 \\
        \midrule
        XGBoost & runs+cons.+var. & yes & \phantom{0}9.13 & 10.18 & \phantom{0}9.44 \\
        XGBoost & runs+cons.+var. & no  & \phantom{0}9.32 & 11.70 & 10.00 \\
        \midrule
        TCN & raw signal & yes & 12.07 & 10.69 & \phantom{0}9.86 \\
        TCN & raw signal & no  & 14.22 & 16.28 & 15.41 \\
        \midrule
        ASC & Beta posterior & --- & 12.82 & 14.56 & 18.33 \\
        \bottomrule
    \end{tabular}
\end{table}

\section{Conclusion and Future Directions}
\label{sec:conclusion}

We present a framework that plugs a Text-to-SQL correctness judge into a convergence-prediction pipeline: from the running consistency of its binary labels, we predict when that consistency has converged and stop sampling early, saving compute and improving accuracy over a fixed-run budget. Beyond classifying \emph{whether} convergence has occurred, our method also detects \emph{when} it first happens, and it outperforms both a statistical (non-learning) Beta-Bernoulli stopping rule and the fixed-run regime. We further use the run autocorrelation to quantify how close the data is to i.i.d., and apply order-permutation augmentation at a strength tuned to that measurement, yielding more useful training data. It also degrades gracefully under label noise (AUC $0.908 \to 0.876$ at $5\%$ flips), making it a promising candidate for production settings where the judge is itself a learned, noisy component.

One natural direction follows: identifying which execution-correctness judge models best fit this framework by studying how each judge's error profile interacts with the convergence signal.

\section*{Limitations}
\label{sec:limitations}

Our labeled-data volume is limited: with more data, deeper sequence models could overtake the feature-based models that dominate this regime. The expected (gold) SQL is written by human annotators, so defining correctness involves subjective judgment and the resulting binary labels are noisy. Finally, all runs use one model (GPT-4.1), so results may not fully transfer to other models or serving conditions.

\bibliography{references}

\appendix
\raggedbottom
\section{Experimental Setup}
\label{sec:experimental_setup}

This appendix provides additional details on our experimental setup, hyperparameter search, and per-dataset performance.

\begin{figure*}[t]
    \centering
    \includegraphics[width=\textwidth]{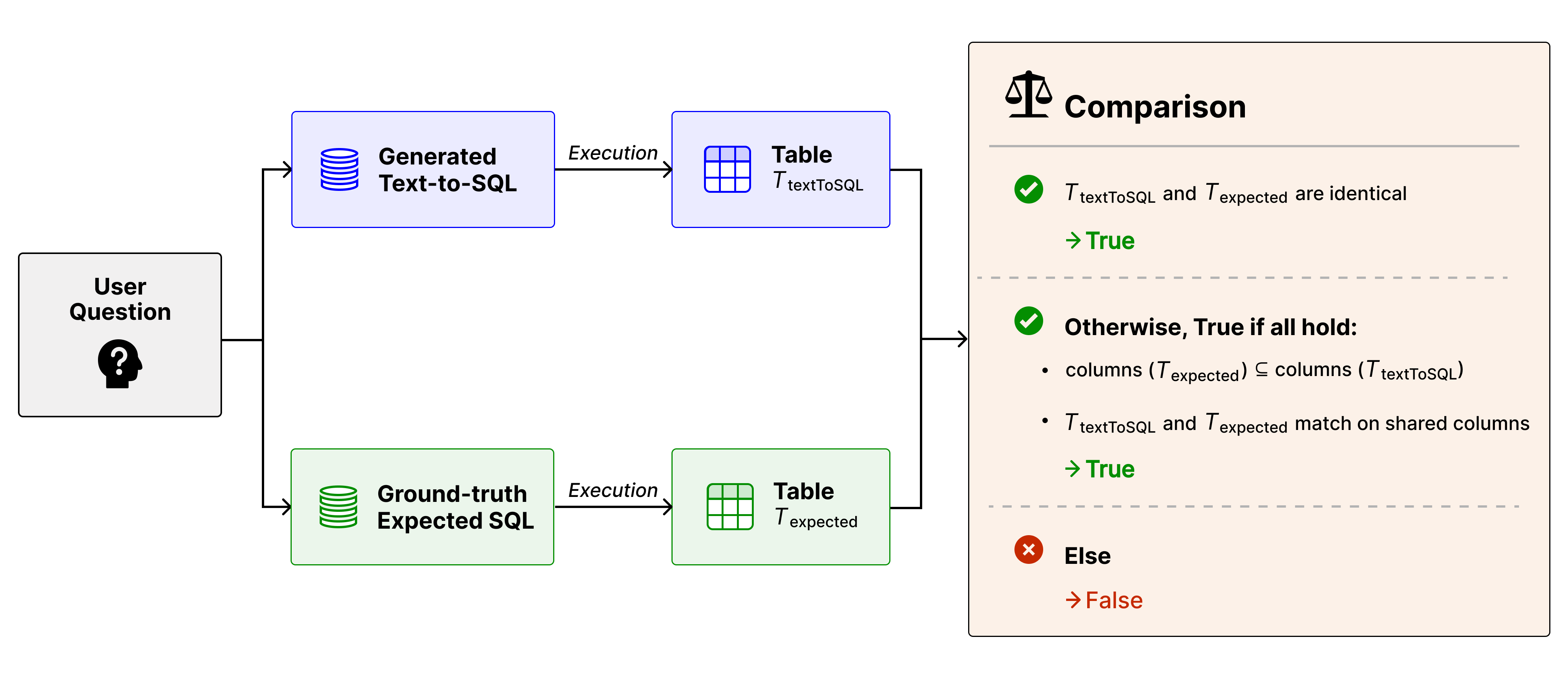}
    \caption{Ground-truth labeling pipeline. For each run, the table executed from the generated SQL is compared against the table executed from the manually annotated expected SQL. Identical tables are labeled \texttt{True}; otherwise, if the generated table has at least as many columns as expected (so all expected information is present), a value-by-value comparison is performed.}
    \label{fig:pipeline}
\end{figure*}

\begin{figure}[t]
    \centering
    \includegraphics[width=\columnwidth]{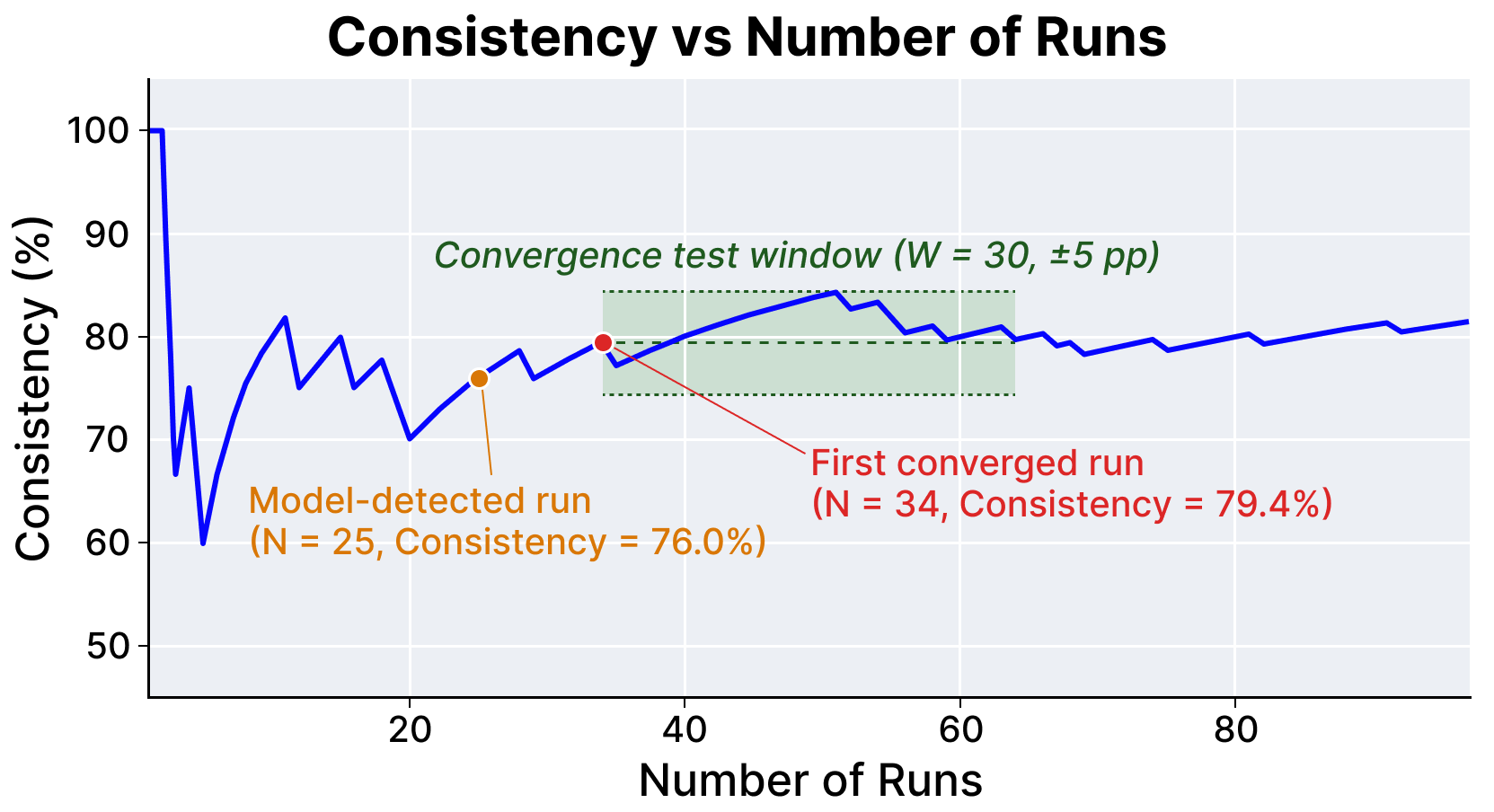}
    \caption{Consistency as a function of the number of runs for an example question. As runs accumulate the consistency settles; our model detects the run at which it has converged and reads off the consistency estimate at that point, close to the first truly converged run. The shaded band marks the $\pm 5$~pp convergence window checked over the following $W=30$ runs.}
    \label{fig:convergence_examples}
\end{figure}

\paragraph{Hyperparameter search.}
For each fold and each model family we ran a grid search over the hyperparameters below and kept the configuration with the best validation performance (ROC AUC for classification, RMSE for detection). For the learned models, the shuffled-data weight $\beta_{\text{shuffle}}$, which trades the natural run order against its permuted copies, is searched jointly with each model's own hyperparameters.
\begin{itemize}
    \item \textbf{Logistic regression:} the regularization parameter \texttt{C} and the elastic-net mixing \texttt{l1\_ratio}, with the penalty fixed to elastic-net and the \texttt{saga} solver.
    \item \textbf{XGBoost:} \texttt{max\_depth}, \texttt{learning\_rate}, \texttt{n\_estimators}, \texttt{min\_child\_weight}, and \texttt{gamma}.
    \item \textbf{1-D TCN:} the architecture is fixed (described below); we search dropout, learning rate, and batch size, with the number of epochs set by early stopping on the validation fold.
    \item \textbf{Beta-Bernoulli baseline:} the credible-interval level and the interval-width threshold at which to stop, with the prior fixed to $\mathrm{Beta}(1,1)$.
\end{itemize}

\paragraph{TCN architecture.}
The raw-signal model is a small fixed network: three dilated 1-D convolutions (kernel $3$, $16$ channels, dilations $1, 2, 4$, a $15$-run receptive field, with ``same'' padding to preserve length), then global average pooling and a linear head to a single sigmoid logit. The pooling makes it length-agnostic, so the same model applies at every run $n$; dropout, tuned on the validation fold, is its main regularizer.

\paragraph{Per-dataset performance.}
Best full-feature configuration per dataset (Table~\ref{tab:best_logistic_by_dataset}), showing performance is consistent across domains: AUC stays in $[0.892, 0.920]$ and RMSE in $[7.63, 8.59]$.

\begin{table}[!htb]
    \centering
    \caption{Per-dataset performance of the best logistic-regression configuration using all features. AUC measures convergence-prediction quality, and RMSE measures the error in detecting the first converged run.}
    \label{tab:best_logistic_by_dataset}
    \footnotesize
    \setlength{\tabcolsep}{4pt}
    \begin{tabular}{lcc}
        \toprule
        \textbf{Dataset} & \textbf{AUC} & \textbf{RMSE} \\
        \midrule
        BIRD         & 0.920 & 8.33 \\
        Dataset A    & 0.911 & 8.59 \\
        Dataset B    & 0.892 & 7.63 \\
        \bottomrule
    \end{tabular}
\end{table}

\paragraph{Feature importance.}
Mean absolute SHAP values~\citep{lundberg2017shap} for the XGBoost full-feature classification model (Table~\ref{tab:shap}); run count and current consistency dominate.

\begin{table}[!htb]
    \centering
    \caption{Mean absolute SHAP values for the XGBoost full-feature convergence-classification model.}
    \label{tab:shap}
    \footnotesize
    \setlength{\tabcolsep}{6pt}
    \begin{tabular}{lc}
        \toprule
        \textbf{Feature} & \textbf{Mean SHAP} \\
        \midrule
        Number of runs ($n$)            & 1.931 \\
        Current consistency $\text{C}(n)$ & 0.906 \\
        Variation $\sigma^2_{30}$       & 0.685 \\
        Variation $\sigma^2_{15}$       & 0.350 \\
        Variation $\sigma^2_{10}$       & 0.107 \\
        Variation $\sigma^2_{5}$        & 0.057 \\
        \bottomrule
    \end{tabular}
\end{table}

\paragraph{Full ablation tables.}
Complete single-feature ablations for classification (Table~\ref{tab:auc_full}) and detection (Table~\ref{tab:rmse_full}).

\begin{table*}[t]
    \centering
    \begin{minipage}[t]{0.48\textwidth}
    \centering
    \caption{Full classification results (ROC AUC), all feature configurations. The $5\%$ and $15\%$ noise columns randomly flip that fraction of the \texttt{True}/\texttt{False} labels.}
    \label{tab:auc_full}
    \footnotesize
    \setlength{\tabcolsep}{3pt}
    \begin{tabular}{llcccc}
        \toprule
        \textbf{Method} & \textbf{Features} & \textbf{Shuf.} & \textbf{clean} & \textbf{5\%} & \textbf{15\%} \\
        \midrule
        Logistic & runs+cons.+var.        & yes & 0.908 & 0.876 & 0.870 \\
        Logistic & runs+cons.+var.        & no  & 0.905 & 0.867 & 0.861 \\
        Logistic & cons. only       & yes & 0.812 & 0.468 & 0.557 \\
        Logistic & cons. only       & no  & 0.826 & 0.563 & 0.557 \\
        Logistic & variation only & yes & 0.784 & 0.642 & 0.668 \\
        Logistic & variation only & no  & 0.812 & 0.612 & 0.660 \\
        Logistic & runs only              & yes & 0.744 & 0.836 & 0.859 \\
        Logistic & runs only              & no  & 0.735 & 0.835 & 0.857 \\
        \midrule
        XGBoost & runs+cons.+var.        & yes & 0.913 & 0.887 & 0.878 \\
        XGBoost & runs+cons.+var.        & no  & 0.899 & 0.882 & 0.856 \\
        XGBoost & cons. only       & yes & 0.741 & 0.718 & 0.683 \\
        XGBoost & cons. only       & no  & 0.723 & 0.730 & 0.664 \\
        XGBoost & variation only & yes & 0.805 & 0.715 & 0.672 \\
        XGBoost & variation only & no  & 0.794 & 0.707 & 0.674 \\
        XGBoost & runs only              & yes & 0.739 & 0.830 & 0.858 \\
        XGBoost & runs only              & no  & 0.725 & 0.828 & 0.853 \\
        \midrule
        TCN & raw signal & yes & 0.881 & 0.862 & 0.861 \\
        TCN & raw signal & no  & 0.760 & 0.837 & 0.860 \\
        \midrule
        ASC & Beta posterior & --- & 0.876 & 0.848 & 0.808 \\
        \bottomrule
    \end{tabular}
    \end{minipage}\hfill
    \begin{minipage}[t]{0.48\textwidth}
    \centering
    \caption{Full detection results (RMSE), all configurations (lower is better). The $5\%$ and $15\%$ noise columns randomly flip that fraction of the \texttt{True}/\texttt{False} labels.}
    \label{tab:rmse_full}
    \footnotesize
    \setlength{\tabcolsep}{3pt}
    \begin{tabular}{llcccc}
        \toprule
        \textbf{Method} & \textbf{Features} & \textbf{Shuf.} & \textbf{clean} & \textbf{5\%} & \textbf{15\%} \\
        \midrule
        Logistic & runs+cons.+var.        & yes & \phantom{0}8.18 & \phantom{0}8.21 & \phantom{0}8.30 \\
        Logistic & runs+cons.+var.        & no  & \phantom{0}9.43 & \phantom{0}8.98 & \phantom{0}8.59 \\
        Logistic & cons. only       & yes & 14.32 & 18.27 & 19.86 \\
        Logistic & cons. only       & no  & 14.32 & 18.22 & 19.76 \\
        Logistic & variation only & yes & 14.32 & 18.42 & 19.37 \\
        Logistic & variation only & no  & 14.32 & 19.34 & 19.36 \\
        Logistic & runs only              & yes & 12.25 & 11.49 & \phantom{0}9.70 \\
        Logistic & runs only              & no  & 14.22 & 13.14 & 10.02 \\
        \midrule
        XGBoost & runs+cons.+var.        & yes & \phantom{0}9.13 & 10.18 & \phantom{0}9.44 \\
        XGBoost & runs+cons.+var.        & no  & \phantom{0}9.32 & 11.70 & 10.00 \\
        XGBoost & cons. only       & yes & 14.32 & 20.60 & 19.63 \\
        XGBoost & cons. only       & no  & 14.32 & 19.66 & 19.45 \\
        XGBoost & variation only & yes & 14.32 & 18.70 & 19.55 \\
        XGBoost & variation only & no  & 14.32 & 19.08 & 19.58 \\
        XGBoost & runs only              & yes & 13.52 & 13.93 & 10.48 \\
        XGBoost & runs only              & no  & 13.39 & 14.20 & 11.03 \\
        \midrule
        TCN & raw signal & yes & 12.07 & 10.69 & \phantom{0}9.86 \\
        TCN & raw signal & no  & 14.22 & 16.28 & 15.41 \\
        \midrule
        ASC & Beta posterior & --- & 12.82 & 14.56 & 18.33 \\
        \bottomrule
    \end{tabular}
    \end{minipage}
\end{table*}

\FloatBarrier
\section{Labeling Examples}
\label{sec:sample_data}

\noindent This appendix shows three illustrative examples from the public BIRD financial database, each with the expected (gold) SQL and a model-generated SQL. The two proprietary customer datasets (A and B) follow a similar relational structure (a small number of joined business tables) and are labeled the same way. A run is labeled \textbf{True} when the generated query's execution result matches the gold result, and \textbf{False} otherwise.

\noindent
\fcolorbox{gray!30}{gray!10}{\parbox{0.94\columnwidth}{%
\textbf{Example 1.}\\[3pt]
\textit{Question:} How many accounts who choose issuance after transaction are staying in the East Bohemia region?\\[4pt]
\textit{Expected SQL:}\\
{\scriptsize\texttt{SELECT COUNT(T2.account\_id)\\
FROM public.district AS T1\\
INNER JOIN public.account AS T2\\
\phantom{xx}ON T1.district\_id = T2.district\_id\\
WHERE T1.A3 = 'east Bohemia'\\
\phantom{xx}AND T2.frequency = 'POPLATEK PO OBRATU'}}\\[4pt]
\textit{Generated SQL:}\\
{\scriptsize\texttt{SELECT COUNT(a.account\_id) AS account\_count\\
FROM public.account AS a\\
INNER JOIN public.district AS d\\
\phantom{xx}ON a.district\_id = d.district\_id\\
WHERE a.frequency = 'POPLATEK PO OBRATU'\\
\phantom{xx}AND d.a3 = 'east Bohemia'}}\\[4pt]
\textit{Result:} both execute to \texttt{13}, \textbf{match (True)}.%
}}
\\[8pt]

\noindent
\fcolorbox{gray!30}{gray!10}{\parbox{0.94\columnwidth}{%
\textbf{Example 2.}\\[3pt]
\textit{Question:} How many customers who choose statement of weekly issuance are Owner?\\[4pt]
\textit{Expected SQL:}\\
{\scriptsize\texttt{SELECT COUNT(T1.account\_id)\\
FROM public.account AS T1\\
INNER JOIN public.disp AS T2\\
\phantom{xx}ON T1.account\_id = T2.account\_id\\
WHERE T2.type = 'OWNER'\\
\phantom{xx}AND T1.frequency = 'POPLATEK TYDNE'}}\\[4pt]
\textit{Generated SQL:}\\
{\scriptsize\texttt{SELECT COUNT(DISTINCT disp.client\_id)\\
\phantom{xxxxxx}AS owner\_count\\
FROM public.disp\\
INNER JOIN public.account\\
\phantom{xx}ON disp.account\_id = account.account\_id\\
WHERE disp.type = 'OWNER'\\
\phantom{xx}AND account.frequency = 'POPLATEK TYDNE'}}\\[4pt]
\textit{Result:} both execute to \texttt{240}, despite the structurally different queries, \textbf{match (True)}.%
}}
\\[8pt]

\noindent
\fcolorbox{gray!30}{gray!10}{\parbox{0.94\columnwidth}{%
\textbf{Example 3.}\\[3pt]
\textit{Question:} What is the gender of the oldest client who opened his/her account in the highest average salary branch?\\[4pt]
\textit{Expected SQL:}\\
{\scriptsize\texttt{SELECT T2.gender\\
FROM public.district AS T1\\
INNER JOIN public.client AS T2\\
\phantom{xx}ON T1.district\_id = T2.district\_id\\
ORDER BY T1.A11 DESC NULLS LAST,\\
\phantom{xxxxxxxx}T2.birth\_date ASC NULLS FIRST\\
LIMIT 1}}\\[4pt]
\textit{Generated SQL:}\\
{\scriptsize\texttt{SELECT client.gender\\
FROM district\\
INNER JOIN account\\
\phantom{xx}ON account.district\_id = district.district\_id\\
INNER JOIN disp\\
\phantom{xx}ON disp.account\_id = account.account\_id\\
INNER JOIN client\\
\phantom{xx}ON disp.client\_id = client.client\_id\\
WHERE district.a11 =\\
\phantom{xxxx}(SELECT MAX(a11) FROM district)\\
ORDER BY client.birth\_date ASC\\
LIMIT 1}}\\[4pt]
\textit{Result:} gold returns \texttt{M}, generated returns \texttt{F}, \textbf{mismatch (False)}.%
}}

\end{document}